\documentclass{article}
\pdfoutput=1
\usepackage{graphicx}
\usepackage{algorithm}
\usepackage{algorithmic}
\usepackage{tabularx} 
\usepackage{newfloat}
\usepackage{listings}
\usepackage{amssymb}
\usepackage{diagbox}
\graphicspath{{./}}
\usepackage{doi}

\usepackage{graphicx}
\usepackage{amsmath}
\usepackage{amssymb}
\usepackage{booktabs}

\usepackage{ragged2e} 

\newcommand{\HRule}[1]{\rule{\linewidth}{#1}}

\begin{document}
	
\title{
	\HRule{2pt} \\
	\textbf{Hierarchical Dynamic Masks for Visual Explanation of Neural Networks}
	\HRule{0.5pt}
}

\author{
	\textbf{Yitao Peng},
	\textbf{Longzhen Yang},
	\textbf{Yihang Liu},
	\textbf{Lianghua He\textsuperscript{*}}\\
	College of Electronic and Information Engineering Tongji University\\
	4800 Cao’an Highway, Shanghai, China 201804\\
	\{pyt, 2111131, yanglongzhen, helianghua\}@tongji.edu.cn
}
\date{}

\maketitle	
	
\begin{abstract}	
	Saliency methods generating visual explanatory maps representing the importance of image pixels for model classification is a popular technique for explaining neural network decisions. Hierarchical dynamic masks (HDM), a novel explanatory maps generation method, is proposed in this paper to enhance the granularity and comprehensiveness of saliency maps. First, we suggest the dynamic masks (DM), which enables multiple small-sized benchmark mask vectors to roughly learn the critical information in the image through an optimization method. Then the benchmark mask vectors guide the learning of large-sized auxiliary mask vectors so that their superimposed mask can accurately learn fine-grained pixel importance information and reduce the sensitivity to adversarial perturbations. In addition, we construct the HDM by concatenating DM modules. These DM modules are used to find and fuse the regions of interest in the remaining neural network classification decisions in the mask image in a learning-based way. Since HDM forces DM to perform importance analysis in different areas, it makes the fused saliency map more comprehensive. The proposed method outperformed previous approaches significantly in terms of recognition and localization capabilities when tested on natural and medical datasets.
\end{abstract}
	
\thispagestyle{empty}
\newpage
\setcounter{page}{2}

\section{Introduction}
Neural networks \cite{huang2017densely,liu2021swin,liu2022convnet} have made remarkable achievements in areas such as image recognition \cite{traore2018deep,dosovitskiy2020image}. So people began to think about the logic of neural network decision-making. Many methods for explaining neural networks have been proposed \cite{patricio2022explainable}. Among them, the method of generating saliency maps to point out the image regions that neural networks pay attention to when making decisions has been widely studied. The generation of saliency maps \cite{olah2018building} is divided into three mainstream methods: perturbation-based methods \cite{petsiuk2018rise,yuan2020interpreting}, gradient-based methods \cite{sundararajan2017axiomatic}, and CAM-based methods \cite{muhammad2020eigen,jiang2021layercam}.

Gradient-based methods \cite{sundararajan2017axiomatic,shrikumar2017learning} propagate the gradient of the target class to the input layer via backpropagation to generate regions that have a large influence on the prediction. But the explanations provided by these methods are less reliable and sensitive to adversarial perturbations \cite{patricio2022explainable}. The method based on the input perturbation \cite{fong2017interpretable,petsiuk2018rise} monitors the change of the prediction probability of a specific class while occluding different image regions, and generates a saliency map describing the importance of the network prediction probability by setting specific perturbation rules and discriminant methods. CAM-based method \cite{zhou2016learning} proposes class activation maps (CAM), which provide visual explanations based on linearly combining activation maps of convolutional neural networks. \cite{chattopadhay2018grad} have removed the requirements of the CAM-based methods for the network structure and improved the accuracy of positioning.

In this paper, in order to alleviate the problem that previous saliency map methods generate interpretation maps with rough localization and cannot express the importance of pixels in a fine-grained manner, we propose DM. It trains mask vectors of different sizes through an optimization-based approach. A single element of a small-sized mask vector corresponds to a large receptive field, and there is less adversarial effect during learning. Large size mask vectors have small receptive fields and can extract fine-grained information from images. In order to combine the advantages of both, we let the small-size mask vector guide the learning of the large-size mask vector, reducing the mask vector's adversarial sensitivity while improving its ability to mine fine-grained information. This enables the DM's final fused masks to generate fine-grained explanatory map.

Additionally, we consider how to more completely capture regions of interest for neural network classification. An image may have multiple regions that contribute to the network's decision, but previous methods often only find one main decision region and ignore others. To avoid this problem, we propose HDM that exploits DM hierarchically. It masks the salient regions found by the DM on the image, re-inputs the masked image into the next DM module to force it to find the regions in the image that the network cares about in the remaining images, and then fuses the above regions according to their importance through a learning-based method. HDM can not only combine the fine search ability of DM, but also conduct a comprehensive search for the decision-making area of the entire image. This makes the saliency maps it generates not only detailed but also comprehensive.

The key contributions of our work are as follows: 

\begin{itemize}
	\item We propose a learning-assisted module DM using vectors of various sizes, which can analyze the importance of image pixels to model classification in detail.
	\item We propose a visual interpretation method HDM that leverages DM hierarchically to generate fine-grained and comprehensive saliency maps.
	\item We conduct multiple experiments on natural and medical datasets and verify that our method achieves state-of-the-art in recognition and localization capabilities.
\end{itemize}

\begin{figure*}[!t]
	\centering
	{\includegraphics[width=1.0\linewidth]{{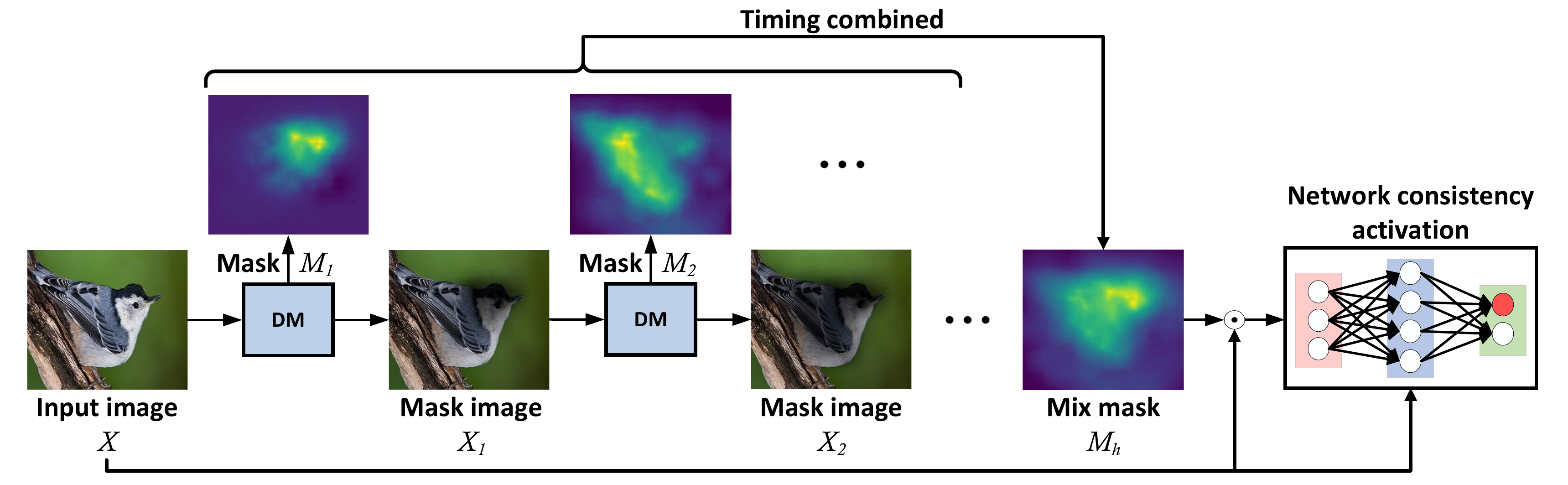}}%
		\label{fig_overall}}
	\caption{Overall architecture of HDM. It is formed by stacking several DM blocks. The original image is input into the DM block to generate a mask and mask image, and then the mask image is continuously input into the next DM block to generate the next mask and mask image. Several masks are obtained by repeating the above operations. All the masks are mixed to generate the final mixed mask through the learning-based timing combination method, and the mixed mask image generated by the dot multiplication of the mixed mask and the original image is trained with the original image to calculate the activation consistency to form the final mixed mask.}
	\label{fig_HDM}
\end{figure*}

\section{Related work}\label{sec:RELATED_WORK}
Existing saliency methods can be divided into three categories, and we briefly introduce these methods in this section.

The first is a perturbation-based method. \cite{fong2017interpretable} use the optimization method to find the areas in the image that are helpful for the network decision-making, and cover the areas in the picture that are not related to the classification. \cite{petsiuk2018rise} detects the importance of various regions in the image in the decision of the network by using a randomly masked version of the input image.

The second is a gradient-based method. Gradient \cite{simonyan2013deep} directly computes the derivative of a class score with respect to an input image using a first-order Taylor expansion. It captures local sensitivity by representing the change at each input location through gradients. \cite{shrikumar2017learning} proposed DeepLIFT to decompose the neural network’s output prediction. Further, \cite{sundararajan2017axiomatic} proposed Integrated Gradients, which uses small changes in the feature space between the interpolated image and the input image to measure the correlation between feature changes and model predictions.

The third is a CAM-based approach.Grad-CAM \cite{selvaraju2017grad} uses the gradient information flowing into the last convolutional layer of a convolutional neural networks to assign importance values to each neuron. In order to increase the positioning accuracy of CAM, \cite{chattopadhay2018grad} proposed Grad-CAM++, which added an additional weight to weigh the elements of the gradient map. FullGrad \cite{srinivas2019full} more fully explains model behavior by aggregating full gradient components. \cite{wang2020score} proposed Score-CAM to solve the problem of easily finding false confidence samples. Ablation-CAM \cite{ramaswamy2020ablation} was present to explore the strength of each factor’s contribution to overall model. Eigen-CAM \cite{muhammad2020eigen} makes CAM generation easier and more intuitive. XGrad-CAM\cite{fu2020axiom} introduces sensitivity and conservation axioms for CAM methods. To generate a more comprehensive saliency map, LayerCAM \cite{jiang2021layercam} considers the role of each spatial location in the feature map using element-level weights.

\begin{figure}[!t]
	\centering
	{\includegraphics[width=1.0\linewidth]{{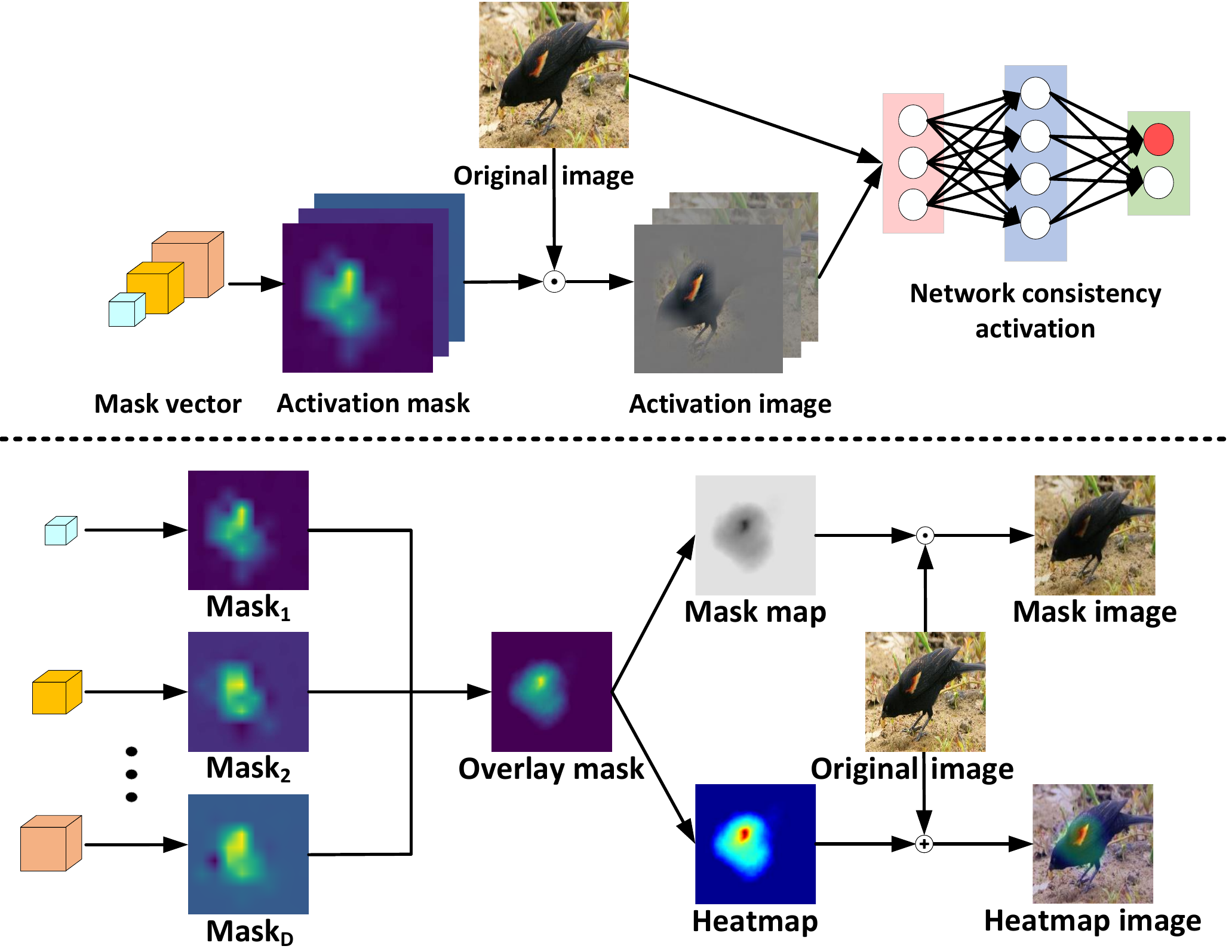}}%
		\caption{The flow of a DM. Above the line, mask vectors of different sizes are trained by constraining the original image and activation image to have consistent activations on nodes of the neural network. The image below the line shows that the upsampled learned mask vectors are stacked to produce an overlay image. The mask image is obtained by multiplying the overlay mask with the original image. The overlay mask uses JET colormap to generate heatmap, which is added to the original image to get heatmap image.}
		\label{fig_dm1}}
\end{figure}

\section{Methodology}\label{sec:Methods}
The overall architecture of HDM and generation method are described in Section \ref{section31}. In addition, Section \ref{section32} introduces DM to fine study the importance of images to network.
\subsection{Hierarchical Masks Generation} \label{section31}
We propose to use DM hierarchical timing to find the regions of the image that neural networks focus on when making decisions. After the current DM finds a decision-making area, it masks this area in the image, and then inputs the masked image to the next DM module to analyze whether DM can find other areas that are helpful for neural network decision-making. Since the decision-making area searched by the previous DM module has been shielded, the next DM module must search in the unshielded area. Through hierarchical method, which forces each DM module to search different regions, it is possible to comprehensively learn the region information in the image that the network pays attention to when making decisions. Figure \ref{fig_HDM} is the overall architecture of HDM, which consists of many DM modules. DM precisely generates the areas of interest to the neural network in the input image. HDM generates comprehensive and refined saliency maps by mixing masks generated by individual DM modules through a learning-based timing combination.

Let the input image be $X \in R^{H \times W \times C}$, the DM module is denoted as $Q( \cdot )$. After inputting image $X_{i-1}$ to DM, DM can calculate the mask $M_{i}$ (where $M_{i} \in R^{H \times W \times 1}$) corresponding to the most critical region for decision-making in image $X_{i}$. We have the following iteration formula, generating $S$ decision masks, $X_{0} = X$, $i \in \{1,2,...,S \}$.
\begin{equation}
M_{i} = Q(X_{i-1})
\end{equation}
\begin{equation}
X_{i} = (1 - N(\sum_{j=1}^{i}M_{j}))X_{0}
\end{equation}
where $N(x) = \frac{x - min(x)}{max(x) - min(x)}$ is a normalization function.
The mix mask $M_{h}$ is generated by mixing $S$ masks $M_{j}$ ($j \in \{1, 2,..., S\}$) through the following timing combination.
\begin{equation} \label{eq_sum}
M_{h} = \frac{\sum_{j=1}^{S} w_{j}M_{j} }{\sum_{j=1}^{S} w_{j}}, w_{j} = \sum_{k=j}^{S} v_{j}^{2}
\end{equation}

Note: $J(M,X)=||f_{p}(MX)-f_{p}(X)||^{2}_{2}$, which represents the consistent squared loss generated by $MX$ and $X$ input neural network $f$ at position $p$. In this paper, we let the position $p$ be the node corresponding to the category predicted by the neural network. $L_{R}(M)=\sum^{H}_{u=1}\sum^{W}_{v=1}\frac{|M_{huv}|}{|HW|}$, which is the regularized value corresponding to mask $M$.
\begin{equation} \label{L_MhX_X}
L(M,X) = J(M,X) + \lambda L_{R}(M)
\end{equation}
where $\lambda$ is a regularization factor. We train the weights of $S$ masks by optimizing the following loss in Equation (\ref{L_MhX_X0}) to obtain the mix mask $M_{h}$ according to Equation (\ref{eq_sum}).
\begin{equation} \label{L_MhX_X0}
v_{1}, v_{2}, ..., v_{S}   \gets \min\limits_{v_{1}, v_{2}, ..., v_{S}} L(M_{h},X_{0})
\end{equation}

Since HDM searches the remaining areas that are conducive to decision-making in the current image hierarchically according to time sequence, $M_{j}$ is no less important than $M_{j+1}$. Therefor, we constrain $w_{j} \geq w_{j+1}$ by Equation (\ref{eq_sum}), $j \in \{1,2,...,S-1 \}$.

\subsection{Dynamic Masks Learning} \label{section32}
DM learns the mask vectors by optimizing the activation consistency between the original image and the mask image and the regularization term of the mask vectors. The higher the mask value of the region that is more important to the classification decision of the neural network, the higher the attention of the neural network to this region (refer to supplementary material for explanation). We first set a small-sized benchmark mask vector to roughly learn the attention of the neural network to each area in the image. The size of the mask vector is inversely proportional to its receptive field, and each element of the small-sized mask vector corresponds to a large receptive field in the image, so that the vector elements can accurately learn the importance of each region in the image.

Benchmark mask vectors $\{d_{i}\}^{D}_{i=1}$, $d_{i} \in R^{a_{i} \times b_{i} \times 1}$, $d_{i}$ are initialized to $\tau$. For any $i,j\in\{1,2,...,D\}$, if $i \neq j$ then $a_{i} \neq a_{j}$ or $b_{i} \neq b_{j}$. Upsample function $g(\cdot, h, w)$ can upsample the original vector to a vector of length $h$ and width $w$. For example, $x \in R^{h_{1} \times w_{1} \times 1}$, and $g(x, h_{2}, w_{2}) \in R^{h_{2} \times w_{2} \times 1}$.

\begin{figure*}[!t]
	\centering
	{\includegraphics[width=1.0\linewidth]{{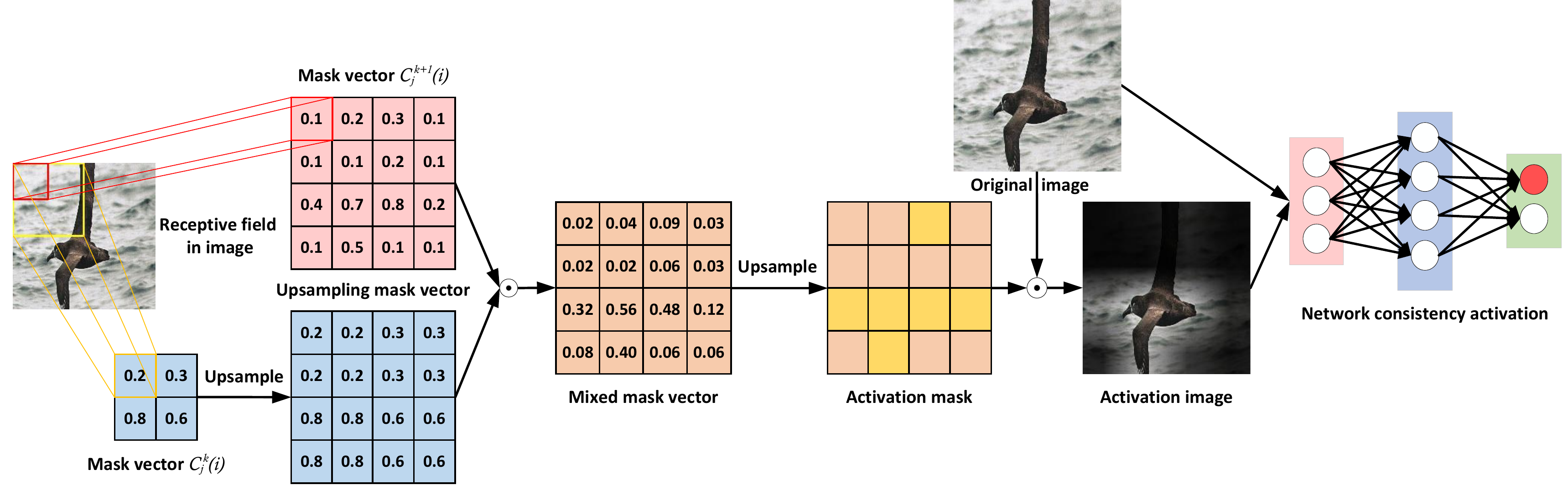}}%
	}
	\caption{An example of training an auxiliary mask vector. The original image size is $224 \times 224$. A feature block of the $2 \times 2$ and $4 \times 4$ mask vector corresponds to the $112 \times 112$ and $56 \times 56$ receptive field of the original image, respectively. The $c^{k+1}_{j}(i)$ to be trained is multiplied element-wise by the upsampled $c^{k}_{j}(i)$ that has been trained, and then the multiplication result is upsampled and multiplied by the original image to generate the activation image. $c^{k+1}_{j}(i)$ is trained by optimizing the consistency activation of the activation and original image.}
	\label{fig_auxiliary}
\end{figure*}

As show in Figure \ref{fig_dm1}, we train $\{d_{i}\}_{i=1}^{D}$ by the activation consistency between the mask image and the original image. Mathematically, we optimize the following loss function.
\begin{equation}
\begin{aligned}
L(g(d_{i}, H, W),X) &  = J(g(d_{i}, H, W), X) \\
&+ \eta_{i} L_{R}(g(d_{i}, H, W))
\end{aligned} 
\end{equation}
where $\eta_{i}$ is a regularization factor. To enable the mask to analyze the importance of region of the image at a fine-grained level, we iteratively generate auxiliary mask vectors $c^{k}_{j}(i)$ on the benchmark mask vectors. $c^{k}_{j}(i)$ is defined as follows:
\begin{equation}
c^{k}_{j}(i) = g(d_{i}, t^{k}_{j}a_{i}, t^{k}_{j}b_{i})
\end{equation}
where $\{t_{j}\}_{j=1}^{T}$ are positive integer hyperparameters, and $k \in \{0, 1, ...., K^{i}_{j} \}$, $K^{i}_{j} = min\{ \left \lfloor \frac{lnH - lna_{i}}{lnt_{j}} \right \rfloor , \left \lfloor \frac{lnW - lnb_{i}}{lnt_{j}} \right \rfloor \}$, $ \left \lfloor \cdot \right \rfloor $ is the floor function. As shown in Figure \ref{fig_auxiliary}, the benchmark mask vector is used as the initial guidance vector, and the auxiliary mask vector of the upper level guides the auxiliary mask vector of the next level to iteratively learn auxiliary vector sets $\{c^{k}_{j}(i)|i\in\{1, ..., D\},j\in \{1, ..., T\}, k \in \{0, ..., K^{i}_{j}\} \}$. The loss function of $c^{k}_{j}(i)$ is as the following Equation (\ref{ckj_optimizer}).
\begin{equation} \label{ckj_optimizer}
\begin{aligned}
&L(g(c^{k}_{j}(i)g(c^{k-1}_{j}(i), t^{k}_{j}a_{i}, t^{k}_{j}b_{i}), H, W), X) \\
& = J(g(c^{k}_{j}(i)g(c^{k-1}_{j}(i), t^{k}_{j}a_{i}, t^{k}_{j}b_{i}), H, W), X) \\
& + \eta_{i}^{(k,j)} L_{R}(g(c^{k}_{j}(i), H, W))
\end{aligned}
\end{equation}

As shown below the line in Figure \ref{fig_dm1}, $M_{c}$ is obtained by stacking the auxiliary masks, and the redundant information of $M_{c}$ is removed to obtain the overlay mask $M_{b}$. $M_{b}$ is the final output result of DM, which can fine-grainedly describe the importance of image regions in neural network decisions.
\begin{equation}
\begin{aligned}
M_{c} = \sum_{i=1}^{D}\sum_{j=1}^{T}\sum_{k=0}^{K^{i}_{j}}c^{k}_{j}(i)
\end{aligned}
\end{equation}
\begin{equation}
\begin{aligned}
M_{b} = N((M_{c} - \gamma)\{M_{c} \geq \gamma\})
\end{aligned}
\end{equation}
where $\gamma$ is the threshold, and $\{ \cdot \}$ represents a truth-valued function, which is 1 if true; otherwise, it equals 0. $N(\cdot)$ is a normalization function.

\section{Experiments}\label{sec:EXPERIMENTAL_STUDIES}
First, the dataset, baseline, and our experimental setup are presented in section \ref{section_data_base} and section \ref{section4.1} for reproducibility. Second, we present the visual interpretation results we generate in section \ref{section4.2}, qualitatively evaluating the performance of our method compared to other previous methods. Furthermore, section \ref{section_heatmap_visualization} visualizes the saliency map generated by the input image in each stage of HDM, showing the flow of HDM for inference. In addition, sections \ref{section4.3} present the results of quantitatively evaluating the fidelity of the interpretation of saliency maps. The quality of the saliency maps are measured by localization ability as described in section \ref{section4.5}. Finally, ablation experiments are performed in section \ref{section_Ablation}.

\begin{figure*}[!t]
	\centering
	{\includegraphics[width=1.0\linewidth]{{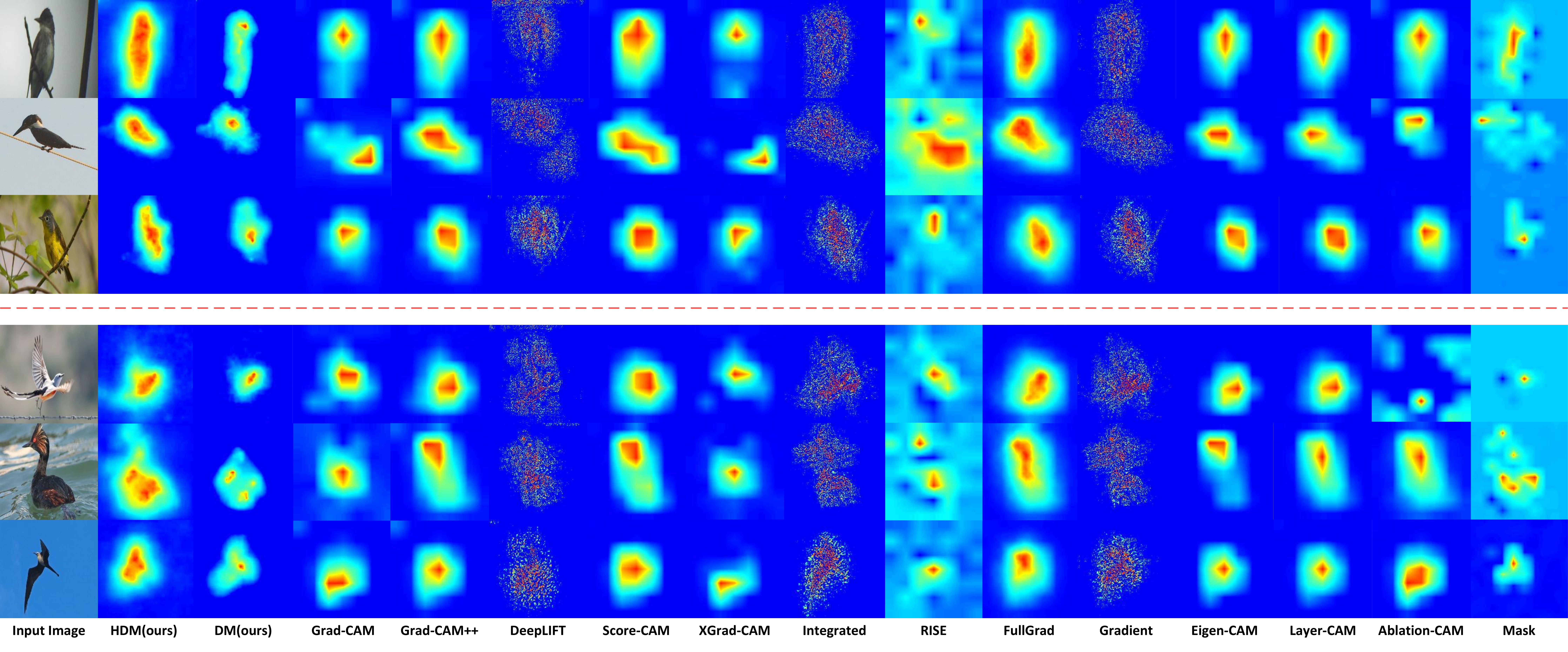}}%
	}
	\caption{Interpretation results for the ResNet50 network using different techniques. The saliency maps generated by each method are normalized to the range [0,1] and visualized using the JET colormap. The saliency map from blue to red indicates a gradual increase in activation probability. Note that the first three rows are examples where the prediction is correct, while the last three rows are wrong.}
	\label{fig_cam}
\end{figure*}

\subsection{Datasets and Baselines} \label{section_data_base}
Experiments were conducted on two public image recognition datasets, including one natural (CUB-200-2011 \cite{wah2011caltech}) and one medical (iChallenge-PM \cite{fu2019palm}) image datasets. We compare 13 state-of-the-art saliency map generation methods. Gradient-based methods: Gradient \cite{simonyan2013deep}, DeepLIFT \cite{shrikumar2017learning}, and Integrated \cite{sundararajan2017axiomatic}. Perturbation-based methods: RISE \cite{petsiuk2018rise}, and Mask \cite{fong2017interpretable}. CAM-based methods: Grad-CAM \cite{selvaraju2017grad}, Grad-CAM++ \cite{chattopadhay2018grad}, Score-CAM \cite{wang2020score}, XGrad-CAM \cite{fu2020axiom}, FullGrad \cite{srinivas2019full}, Eigen-CAM \cite{muhammad2020eigen}, Ablation-CAM \cite{ramaswamy2020ablation}, and Layer-CAM \cite{jiang2021layercam}. The saliency maps generated by the above-mentioned methods are taken as baselines.

\subsection{Experimental Details} \label{section4.1}
In this work, ResNet50 \cite{he2016deep} was chosen as the neural network for saliency maps interpretation. All methods use a ResNet50 model with the same parameters and fixed for visual evaluation. The input image is resized to $224 \times 224 \times 3$, and then normalized using mean vector (0.485, 0.456, 0.406) and variance vector (0.229, 0.224, 0.225). Our training set contains only images and their class labels. ResNet50 is pre-trained on ImageNet \cite{deng2009imagenet} before being trained on the natural and medical images datasets. Each image is trained for $800$ epochs with a learning rate set to $1e-2$. We set $\lambda = 1e-4 $, $\eta_{i} = \eta_{i}^{(k,j)} = 100$. In the CUB-200-2011, we set $d_{i} = i + 5$, $\{d_{i}\}_{i=1}^{6}$, $\{t_{j}\}_{j=1}^{3} = \{2, 3, 5\}$; $\gamma$ is set to the pixel value of the top $25\%$ of the pixel value of the salient image; HDM consists of $3$ DM modules. In the iChallenge-PM, we set $d_{i} = i + 5$, $\{d_{i}\}_{i=1}^{4}$, $\{t_{j}\}_{j=1}^{2} = \{2, 3\}$; $\gamma$ is set to the pixel value of the top $30\%$ of the pixel value of the salient image; HDM consists of $1$ DM modules. The learning-based mix mask iteration $400$ epochs, and its learning rate is $1e-1$.

\begin{figure}[!t]
	\centering
	{\includegraphics[width=1.0\linewidth]{{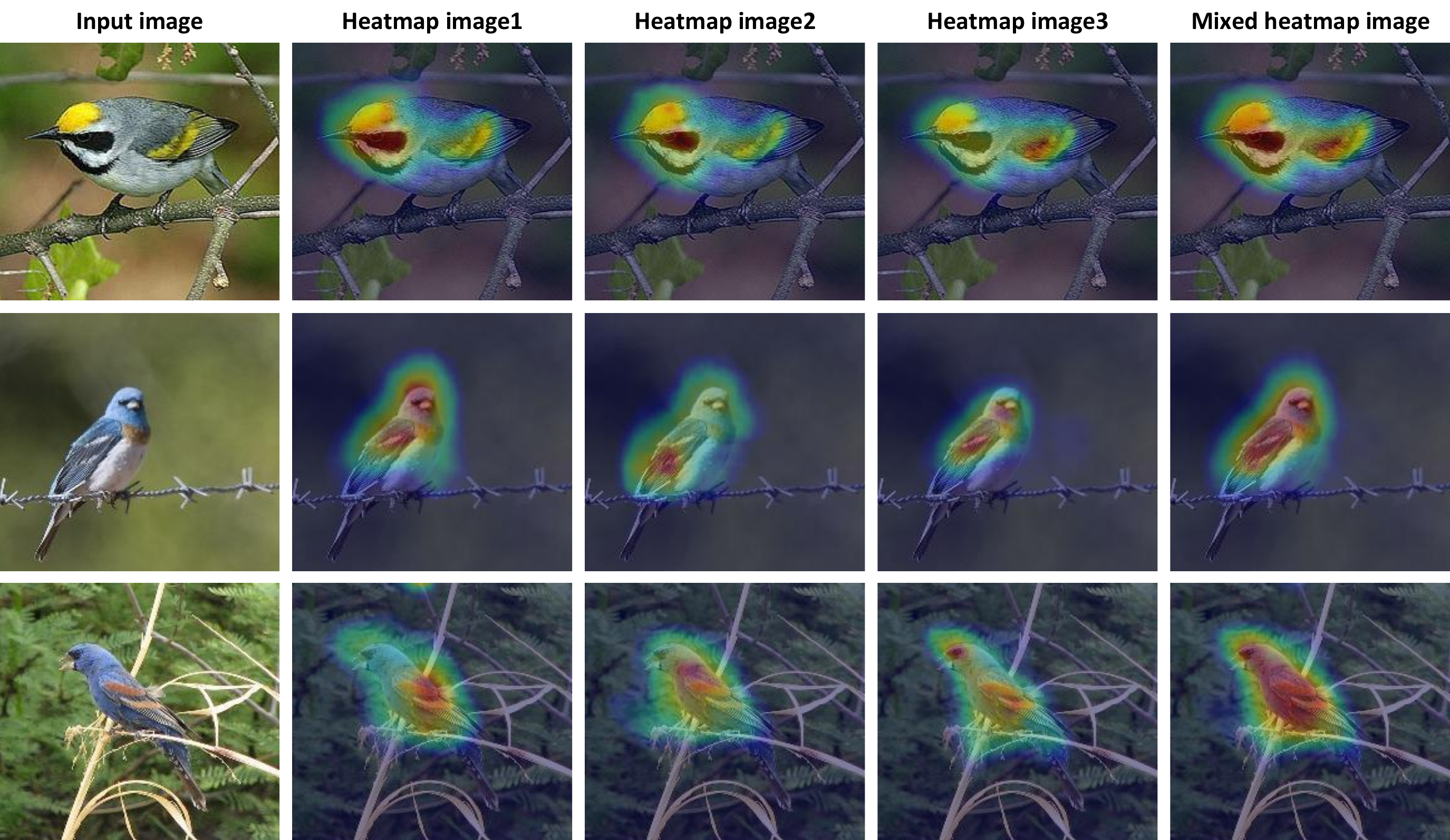}}%
		\caption{The heatmap generated by the whole process of HDM. The first column are the input images; the second, third, and fourth column images are the heatmap images generated after passing through the DM module for the first, second, and third times respectively; the fifth column of images is the heatmap image generated by mixing the second, third, and fourth columns of heatmaps through a learning-based method, which is also the output of HDM.}
		\label{fig_dm123}}
\end{figure}

\subsection{Qualitative Evaluations} \label{section4.2}
Experiments are designed to qualitatively compare the performance of saliency maps generated by different methods. As shown in Figure \ref{fig_cam}, saliency maps point out regions of interest that networks interpret when making decisions, and we compare 13 state-of-the-art saliency map generation methods.

Interpretation results for six different images are reported in Figure \ref{fig_cam}. In each row, the leftmost image is the original image, and the second and third rows are the saliency maps generated by our proposed HDM and DM methods, respectively. The subsequent columns represent the results obtained using different methods. The color of the saliency maps from red to blue represents indicates the gradual decrease in the probability of the network's attention to the region.

As shown in Figure \ref{fig_cam}, the saliency maps generated by CAM-based methods usually only point out the areas of the image that a network pays attention to when making decisions. The form of this region is almost always with a center as the highest activation, diffuses outward and gradually weakens irregularly, so the generated The saliency map has a single structure, lacks integrity, and is not easy for humans to understand. In contrast, the saliency map generated by HDM can discover multiple decision-making regions and integrate them completely. It finely includes the body regions of the bird used for decision making, outlining the silhouette of the bird. The HDM method can completely and accurately generate the areas of concern when making network decisions, which is more comprehensive and easy to understand than the CAM-based method.

The saliency map generated based on the gradient method can roughly describe the contours of the objects concerned by the network decision-making, but the pixels of the saliency map are discrete and sharp, and it is difficult for people to observe which are the key decision-making areas. By comparison, we can see that the saliency map generated by HDM can better describe the outline of the object while retaining the smooth change of activation, which makes it easier for people to observe the importance of each region.

The saliency map generated by the perturbation-based method can mark multiple regions for decision-making, but such saliency maps cover too large areas and have more noise. In contrast, HDM can cover multiple decision regions while paying little attention to regions outside the decision region (bird body), generating saliency maps with little noise.

\begin{table}
	\centering
	\begin{tabular}{lllll}
		\hline
		Evaluation & \multicolumn{2}{l}{Average Drop} & \multicolumn{2}{l}{Average Increase}\\
		\hline
		Percentile & 80\% & 70\% & 80\% & 70\% \\
		\hline
		Mask & 39.1\% & 24.1\% & 15.3\% & 22.3\% \\
		RISE & 31.2\% & 17.4\% & 24.2\% & 35.4\% \\
		Gradient & 97.8\% & 97.4\% & 0.4\% & 0.8\% \\
		DeepLIFT & 97.7\% & 97.5\% & 0.3\% & 0.6\% \\
		Integrated & 98.4\% & 97.8\% & 0.3\% & 0.5\% \\
		Grad-CAM & 31.5\% & 17.0\% & 21.3\% & 35.7\% \\
		Grad-CAM++ & 35.2\% & 19.1\% & 18.2\% & 32.2\% \\
		Score-CAM & 31.2\% & 18.2\% & 19.7\% & 35.3\% \\
		Ablation-CAM & 33.7\% & 19.7\% & 17.3\% & 34.4\% \\
		XGrad-CAM & 33.9\% & 18.7\% & 18.2\% & 36.6\% \\
		Eigen-CAM & 38.5\% & 22.9\% & 17.1\% & 31.8\% \\
		Layer-CAM & 36.6\% & 19.3\% & 16.3\% & 29.6\% \\
		FullGrad & 35.5\% & 19.5\% & 20.1\% & 30.9\% \\
		DM(ours) & 32.6\% & 15.7\% & 26.4\% & 36.9\% \\
		HDM(ours) & \textbf{30.1\%} & \textbf{13.3\%} & \textbf{26.7\%} & \textbf{38.8\%} \\
		\hline
	\end{tabular}
	\caption{Evaluated results on average drop (lower is better) and average increase (higher is better).}
	\label{ADAI}
\end{table}

Overall, our method can generate refined and complete saliency maps, which are less noisy and easy to understand.

\subsection{Processes Visualization} \label{section_heatmap_visualization}
As shown in Figure \ref{fig_dm123}, we show all saliency maps generated by HDM during inference on images from the CUB-200-2011 dataset. When researching on the CUB-200-2011 dataset, we set HDM to consist of 3 DM modules. The first column in Figure \ref{fig_dm123} is the original input image. The second, third, and fourth columns in Figure \ref{fig_dm123} represent the saliency images of the current most-attended regions of the neural network found after inputting the original image or the mask image into the DM module in the three stages, respectively. It can be seen from Figure \ref{fig_dm123} that each DM module can accurately and fine-grainedly find the remaining regions in the current picture that are helpful for neural network decision-making, and the regions of interest found in each stage are almost disjoint. This is in line with HDM's hierarchical search strategy. The mixed heatmap images in the last column are the result of mixing the heatmap images generated by the first three DM modules via a learning-based approach, which makes the final generated saliency maps comprehensive. It can be seen from Figure \ref{fig_dm123} that the final hybrid map generated by HDM is not only comprehensive, but also fine-grained.

\subsection{Faithfulness Evaluation via Image Recognition} \label{section4.3}
The experiment evaluates the fidelity of the interpretation of the saliency map generated by the HDM in the object recognition task, and the average drop and average increase \cite{chattopadhay2018grad} are used as evaluation indicators. This experiment sets the activation value of a specific percentage of all pixels in the saliency map as the threshold value, mutes the pixels below the threshold value, and retains the pixel value above the threshold value to generate an interpretation map (in the experiment, set 70\% and 80\% of the pixels are muted). The average drop is expressed as $\sum_{i=1}^{N} \frac{max(0, Y_{i} - O_{i})}{Y_{i}} \times 100 $, and the average increase is expressed as $\sum_{i=1}^{N} \frac{Sign(Y_{i} < O_{i})}{N}$, where $Y_{i}^{c}$ is the predicted score of image $i$ in category $c$, and $O_{i}^{c}$ is the predicted score of category $c$ with the explanatory map as input. $Sign( \cdot )$ is an indicator function that returns 1 if the input is True. In the experiment, 10 images are randomly selected for each category on CUB-200-2011, and a total of 2000 images are formed for testing. The results are reported in Table \ref{ADAI}. When the threshold is set to occlude 70\% and 80\% of the pixels, HDM achieves 13.3\% and 30.1\% average drop and 38.8\% and 26.7\% average increase, respectively. HDM outperforms other saliency methods. This task shows that HDM can find the most distinguishable regions of the target object as much as possible, and can eliminate as much as possible the regions irrelevant to the target object distinction.

\begin{table}
	\centering
	\begin{tabular}{lcc}
		\toprule
		Evaluation  & Deletion Score & Insertion Score \\
		\midrule
		Mask & 0.0516 & 0.7579 \\
		Grad-CAM & 0.0537 & 0.7939 \\
		Grad-CAM++ & 0.0578 & 0.7937 \\
		Score-CAM & 0.0576 & 0.8076 \\
		Ablation-CAM & 0.0643 & 0.7724 \\
		XGrad-CAM & 0.0582 & 0.7983 \\
		Eigen-CAM & 0.0763 & 0.7751 \\
		Layer-CAM & 0.0601 & 0.7954 \\
		FullGrad & 0.0537 & 0.7958 \\
		DM(ours) & 0.0461 & 0.8171 \\
		HDM(ours) & \textbf{0.0447} & \textbf{0.8189} \\
		\bottomrule
	\end{tabular}
	\caption{Evaluated results on delecion (lower is better) and insertion (higher is better) scores.}
	\label{deletionandinsertion}
\end{table}

\begin{figure}[!t]
	\centering
	{\includegraphics[width=1.0\linewidth]{{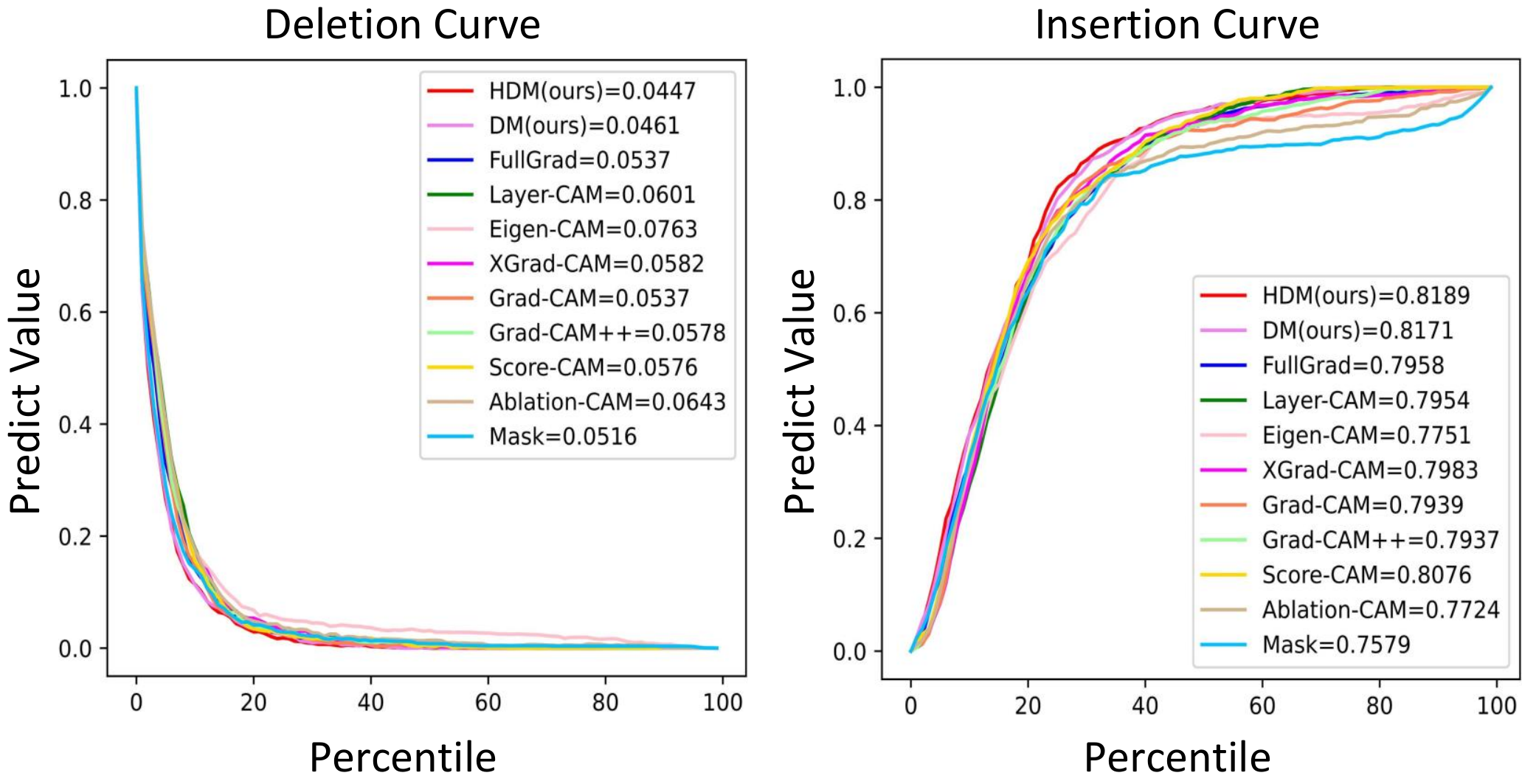}}%
		\caption{Deletion and Insertion curves of the above methods.}
		\label{fig_deletion}}
\end{figure}

To further evaluate the discriminative ability of HDM, we test on the insertion and deletion metrics proposed in \cite{petsiuk2018rise}. Deletion and insertion scores are the areas under the probability curves described by the predicted probability results for deleted or inserted images generated by deleting and inserting pixels from the original image in descending order of saliency map activation values, respectively. This experiment adopts the setting in \cite{wang2020score}, the step size is 1\%, and the pixel value is set to 0 or 1 to remove or introduce pixels. We compare the performance of the methods in Table \ref{deletionandinsertion} on the above-mentioned evaluation metrics. Here we do not compare other perturbation-based methods, since they can produce adversarial effects \cite{petsiuk2018rise} that interfere with these evaluation metric performance. As shown in Figure \ref{fig_deletion}, the deletion and insertion curves produced by each model are displayed. The HDM method achieves the largest AUC and the steepest curve, which shows that the HDM saliency map can best find the area in the current image that has the greatest influence on the network decision. Table \ref{deletionandinsertion} reports the average results for the above-mentioned 2000 images, and our method achieves state-of-the-art on both metrics compared to the respective previous methods.

\begin{table}
	\centering
	\begin{tabular}{lcc}
		\hline
		Evaluation & \multicolumn{2}{c}{Proportion} \\
		\hline
		Dataset & CUB-200-2011  & iChallenge-PM \\
		\hline
		Mask & 33.31\% & 27.31\% \\
		RISE & 29.47\% & 14.33\% \\
		Gradient & 31.58\% & 13.57\% \\
		DeepLIFT & 29.86\% & 13.63\% \\
		Integrated & 31.50\% & 13.59\% \\
		Grad-CAM & 50.34\% & 20.84\% \\
		Grad-CAM++ & 51.37\% & 17.36\% \\
		Score-CAM & 54.52\% & 24.71\% \\
		Ablation-CAM & 48.33\% & 18.14\% \\
		XGrad-CAM & 48.92\% & 18.91\% \\
		Eigen-CAM & 57.41\% & 21.64\% \\
		Layer-CAM & 56.24\% & 20.30\% \\
		FullGrad & 47.61\% & 17.94\% \\
		DM(ours) & 53.81\% & 32.81\% \\
		HDM(ours) & \textbf{57.97\%} & \textbf{32.81\%}\\
		\hline
	\end{tabular}
	\caption{On CUB-200-2011 and iChallenge-PM datasets, the evaluation results of each method in proportion.}
	\label{Percentile}
\end{table}

\subsection{Localization Evaluation} \label{section4.5}
In this section, the localization ability is employed to measure the quality of the generated saliency maps. In the same way as in \cite{wang2020score}, calculate how much energy of the saliency map falls into the segmented foreground region of the target object. Binarize the segmented foreground image given in the test set, assign the foreground area as 1, and assign the background area as 0, multiply the saliency map calculated by each method with the binarized image point by point and sum how much energy is obtained in the target foreground. The evaluation metric Proportion is expressed as $\frac{\sum L^{c}_{(i,j) \in bbox}}{\sum L^{c}_{(i,j) \in bbox} + \sum L^{c}_{(i,j) \notin bbox}}$. We test the localization ability on the CUB-200-2011 and iChallenge-PM datasets, setting the foreground region of the image segmentation label as the bbox region. In the test set of CUB-200-211, 10 images are randomly selected from each category, and a total of 2000 images are formed for experiments; in the test set of iChallenge-PM, 200 images are randomly selected as experiments. The saliency maps generated by HDM have 57.97\% and 32.81\% energy falling in the ground-truth foreground region of the target in CUB-200-2011 and iChallenge-PM, respectively. This verifies that HDM can effectively reduce noisy regions in saliency maps that are not relevant to decisions. Our method achieves state-of-the-art performance in both natural and medical image localization ability.

\begin{figure}[!t]
	\centering
	{\includegraphics[width=1.0\linewidth]{{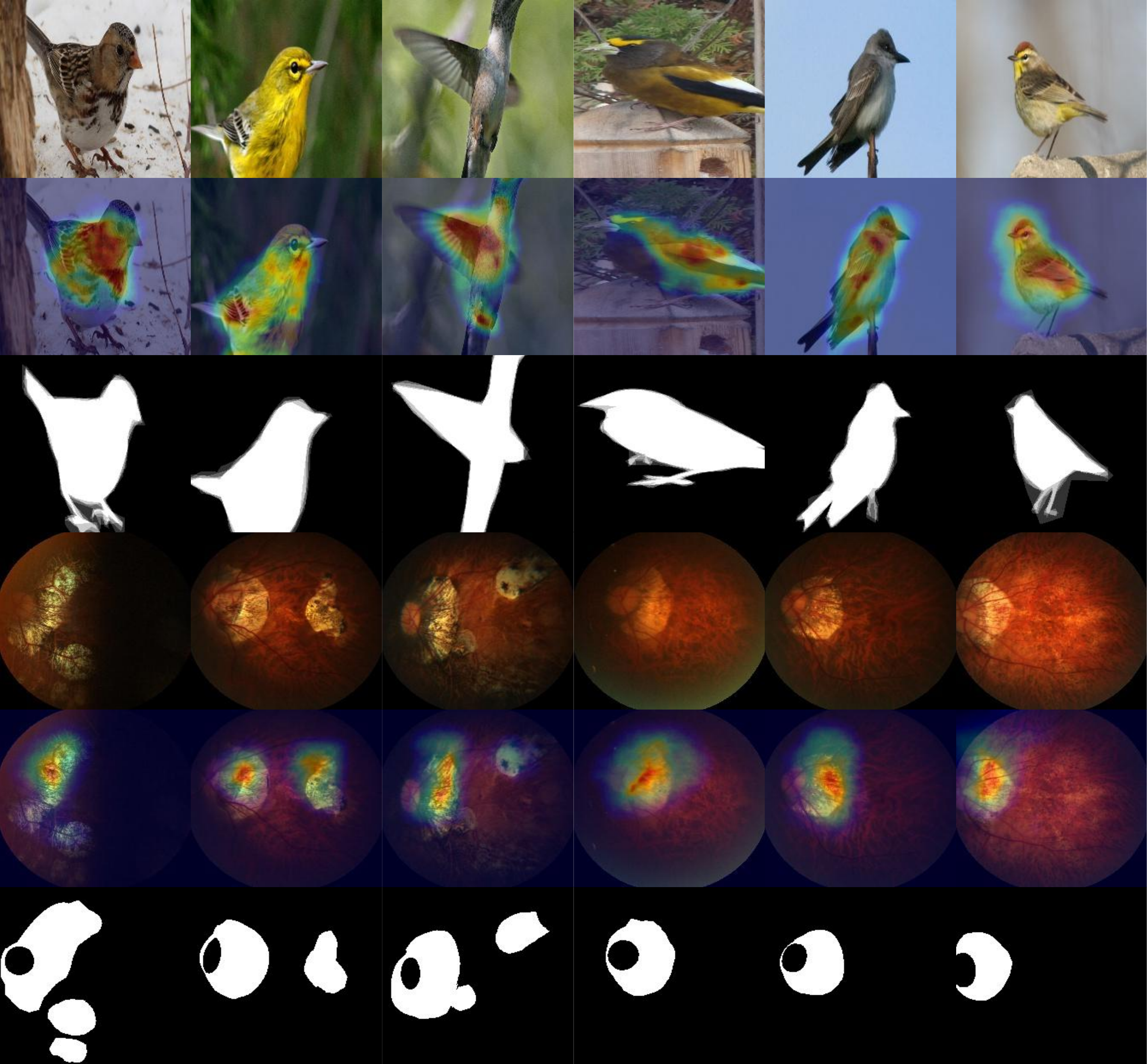}}%
		\caption{Results of saliency maps generated by HDM on bird and fundus images. The first and fourth rows are the original images; the second and fourth rows are the mixture of the saliency map and the original image; the third and sixth rows are the ground-truth.}
		\label{fig_prop_vis}}
\end{figure}

As shown in Figure \ref{fig_prop_vis}, we visualize the results of the saliency maps generated by the HDM on bird and fundus datasets, respectively. In the bird image, the decision-making region found in the saliency map of HDM wraps most of the bird's body, and carefully evaluates the importance of each position on the body. This verifies that HDM is capable of comprehensively locating multiple decision regions in natural images and fine-grained division of the importance of each region. In the fundus images, HDM can also accurately and completely locate pathological areas in the images. The above experiments show that HDM can achieve good localization performance in both natural and medical images.

\subsection{Ablation Study} \label{section_Ablation}
In this section, we compare the performance of saliency maps generated using the DM module alone and HDM consisting of multiple DM hierarchical timings. It can be seen from Table \ref{ADAI} and Table \ref{deletionandinsertion} that the saliency map generated by DM is superior to all previous methods in the evaluation index of recognition ability; Table \ref{Percentile} shows that the localization ability of DM on natural images is weaker than some saliency methods, and the localization ability on medical images reaches the best. From the above-mentioned tables, it can be seen that the hierarchical stacking DM method of HDM can improve the performance of the saliency map on the above indicators. Because DM only focuses on finding the area that a neural network pays attention to when making decisions, but ignores the fact that there may be multiple areas that have an impact on the classification of the neural network. HDM makes the saliency map more comprehensive by using DM multiple times to search for decision regions. Therefore, both the DM module and the method of hierarchically stacking the DM modules are useful and necessary.

\section{Conclusion}
In this paper, we propose HDM to generate refined and comprehensive visual explanatory maps. The DM module is able to generate finely localized saliency maps using masks of various sizes that cooperate with each other. HDM analyzes the masked image hierarchically using the DM module to find areas of concern for multiple neural network decision-making classification decisions in the image, and combines them well into a comprehensive visual interpretation map through a learning-based fusion method. Our method achieves state-of-the-art performance in recognition and localization capabilities on natural images and medical images, and quantitative evaluation verifies that our method can generate more fine-grained and comprehensive saliency maps.

\bibliographystyle{named}
\bibliography{hdm}

\end{document}